\documentclass[a4paper,conference]{IEEEtranx}
\usepackage{etoolbox}
\makeatletter
\usepackage{etoolbox}
\makeatletter
\patchcmd{\@makecaption}
  {\scshape}
  {}
  {}
  {}
\makeatletter
\patchcmd{\@makecaption}
  {\centering}
  {}
  {}
  {}
\makeatletter
\patchcmd{\@makecaption}
  {\\}
  {.\ }
  {}
  {}
\makeatother

\def\tablename{Table}
\def\figurename{Figure}

\ifCLASSINFOpdf
  \usepackage[pdftex]{graphicx}
\else
  \usepackage[dvips]{graphicx}
\fi
\usepackage{xcolor}
\usepackage{amsmath}
\usepackage{amsfonts}
\usepackage{amssymb}
\usepackage{enumitem}
\usepackage{afterpage,array,rotating}
\usepackage{multirow}
\newcolumntype{C}{>{\centering\arraybackslash}m{1.3cm}}
\newcolumntype{D}{>{\centering\arraybackslash}m{1.2cm}}
\def\etal#1{#1} %et al.,}

\def\proc{\em Proc.~}
\def\jornal{\em In~}

\def\fig#1#2#3{
\begin{figure}[!t]
\centering
\includegraphics[width=#3\hsize]{fig/#1}
\caption{#2}
\label{#1}
\end{figure}
}
\def\figw#1#2#3{
\begin{figure*}[!t]
\centering
\includegraphics[width=#3\hsize]{fig/#1}
\caption{#2}
\label{#1}
\end{figure*}
}
\def\tableA{
\begin{table*}[!t]
\caption{Performance comparison on four datasets. Classification accuracies (\%) for each dataset with two types of network are shown.}
\begin{center}
{\normalsize
\begin{tabular}{|l|C|C|C|C|C|C|C|C|}
\hline
\multirow{2}{*}{Method} & \multicolumn{2}{c|}{Cifar-10} & \multicolumn{2}{c|}{Cifar-100} & \multicolumn{2}{c|}{Omniglot} & \multicolumn{2}{c|}{DTD}\\
\cline{2-9}
 & ResNet50 & $\hspace{-3pt}\mbox{ResNet152}$ & ResNet50 & $\hspace{-3pt}\mbox{ResNet152}$ & ResNet50 & $\hspace{-3pt}\mbox{ResNet152}$ & ResNet50 & $\hspace{-3pt}\mbox{ResNet152}$\\
\hline
Normal initialization & 92.62 & 93.47 & 75.16 & 75.59 &\ \ 2.66 &\ 2.37 & 13.68 & \ 5.24\\
Xavier initialization \cite{Xavier_2010} & 92.30 & 93.58 & 73.85 & 75.14 &\ \  5.88 &\ 5.57 & 27.51 & 24.75\\
He initialization \cite{He_2015} & 93.50 & 93.43 & 74.17 & 75.73 &\ \ 4.61 & \ 3.06 & 24.31 & 20.14\\
Proposed method & {\bf 93.76} & {\bf 94.27} & {\bf 77.42} & {\bf 78.21} & {\bf 17.54} & {\bf 18.71} & {\bf 55.03} & {\bf 54.18}\\
\hline
\end{tabular}
}
\label{tableA}
\end{center}
\end{table*}
}
\def\tableB{
\begin{table}[!t]
\caption{Experimental results obtained using various values for hyperparameters in noise data generation. Accuracies for the Cifar-100 dataset are shown.
Network: backbone network (ResNet50 or ResNet152).
\# categories: number of Perlin noise categoreis; $N \times M$ categoreis are generated.
\# instances: number of instances per category.}
\begin{center}
{\normalsize
\begin{tabular}{|l|c|D|D|D|}
\hline
\multirow{2}{*}{Network} & \multirow{2}{*}{\#categories} & \multicolumn{3}{c|}{\#instances}\\\cline{3-5}
& & $100$ & $500$ & $1,000$ \\
\hline\hline
& $10 \times 10$ & 76.28 & 75.99 & 76.44\\
ResNet50 & $18 \times 18$ & 76.95 & 76.92 & 75.53 \\
& $36 \times 36$ & 75.77 & 77.42 & 78.03 \\
\hline\hline
& $10 \times 10$ & 77.09 & 77.09 & 77.77\\
ResNet152 & $18 \times 18$ & 77.92 & 77.69 & 77.23 \\
& $36 \times 36$ & 78.30 & 78.21 & 77.51 \\
\hline
\end{tabular}
}
\label{tableB}
\end{center}
\end{table}
}
\begin{document}
\title{
\vspace{25pt}
{\Large \bf
Initialization Using Perlin Noise for Training
Networks\\
\vspace{-9pt}
with a Limited Amount of Data
}
\vspace{3pt}
}
\author{
\IEEEauthorblockN{Nakamasa Inoue$^{1*}$, Eisuke Yamagata$^{1*}$, Hirokatsu Kataoka$^{2}$
}
\IEEEauthorblockA{
$^{1}$Tokyo Institute of Technology, Tokyo, Japan.\\
$^{2}$National Institute of Advanced Industrial Science and Technology (AIST), Tsukuba, Japan.\\
E-mail: inoue@c.titech.ac.jp, yamagata.e.ab@m.titech.ac.jp, hirokatsu.kataoka@aist.go.jp
\vspace{10pt}
}
}

\maketitle

\begin{abstract}
We propose a novel network initialization method using Perlin noise for training image classification networks with a limited amount of data.
Our main idea is to initialize the network parameters by solving an artificial noise classification problem, where
the aim is to classify Perlin noise samples into their noise categories.
Specifically, the proposed method consists of two steps.
First, it generates Perlin noise samples with category labels defined based on noise complexity.
Second, it solves a classification problem, in which network parameters are optimized to classify the generated noise samples.
This method produces a reasonable set of initial weights (filters) for image classification.
To the best of our knowledge, this is the first work to initialize networks by solving an artificial optimization problem without using any real-world images.
Our experiments show that the proposed method outperforms conventional initialization methods on four image classification datasets.
\end{abstract}

\IEEEpeerreviewmaketitle
    \renewcommand{\footnoterule}{
     \vspace{0pt}
     \noindent\rule{0.48\textwidth}{0.4pt}
     \vspace{0pt}
    }
\renewcommand{\thefootnote}{\fnsymbol{footnote}}
\footnote[0]{* indicates equal contribution.}
\renewcommand{\thefootnote}{\arabic{footnote}}

\section{Introduction}
Image classification is one of the most important topics in the field of pattern recognition and computer vision, with wide applications such as internet search engines, robotics, and security.
Over the last 10 years, many studies have shown the effectiveness of deep neural networks for various image classification tasks, including object recognition and action recognition.
Most neural networks utilize a large-scale dataset for training because network performance increases with dataset size.
For example, deep convolutional networks trained on the ImageNet dataset \cite{Russakovsky_2015}, which consists of 1.2 million images, have been shown to achieve human-level performance in terms of 1,000-class object classification accuracy.

Training networks with a limited amount of data is a challenging problem because network optimization methods often fall into a local solution if the network has many parameters.
A recent trend in attempts to solve this problem is to apply fine-tuning, i.e., to resume training from a pre-trained network.
For example, for object recognition, networks are often first trained on the ImageNet dataset, and then fine-tuned on another small dataset.
This approach works well because ImageNet is large enough to obtain reasonable image representations with some visual filters in hidden layers.
However, its application is often limited to non-commercial use such as research and educational purposes.
In practice, it is not always easy to collect such a large number of images.

\fig{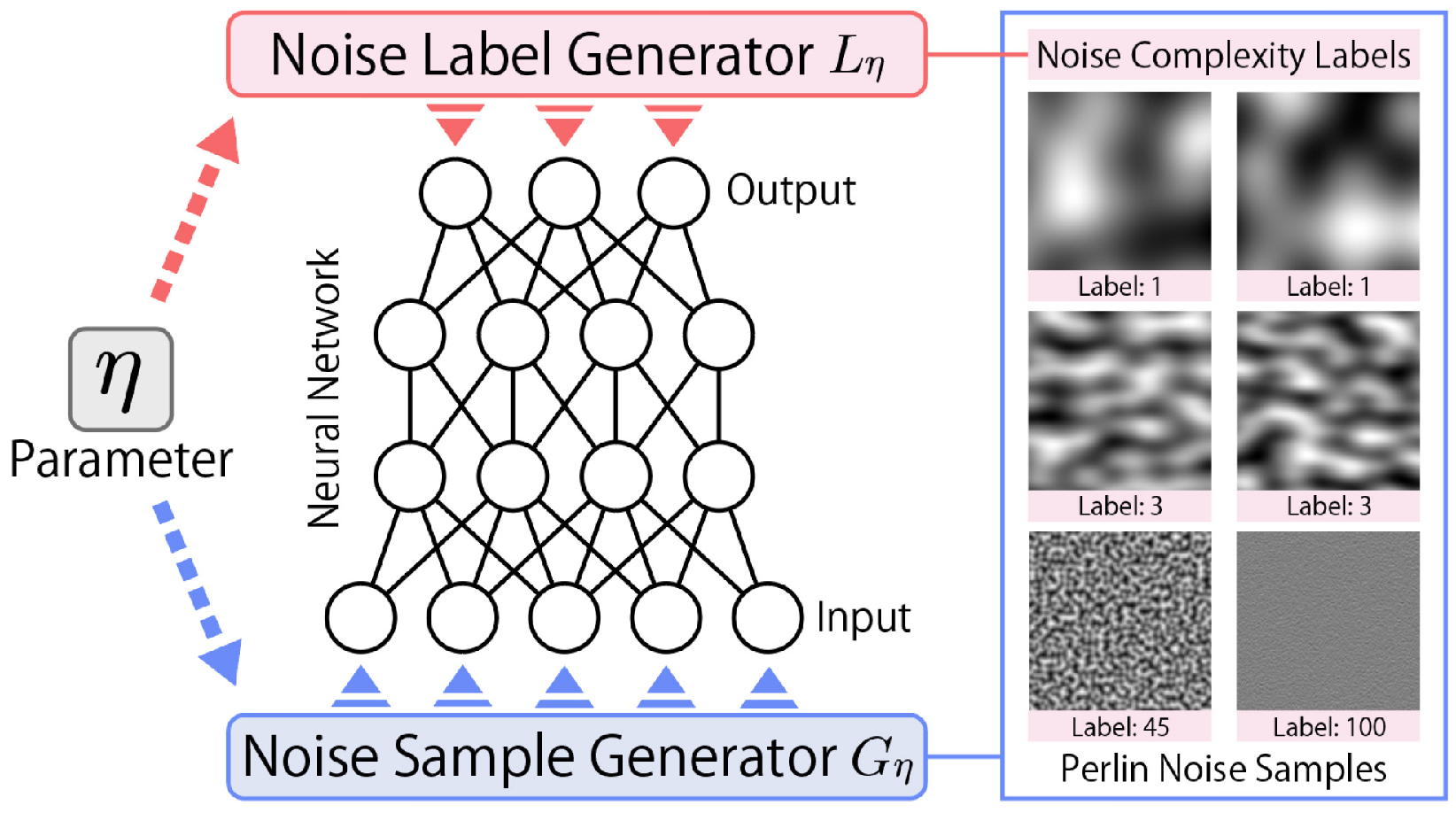}{
{\bf Network initialization using Perlin noise}.
The proposed method consists of two components, namely
a noise sample generator $G_{\eta}$ for generating Perlin noise samples and
a noise label generator $L_{\eta}$ for generating noise category labels.
A network is initialized by solving an artificial noise classification problem, where the
aim is to classify the generated noise samples into their categories.
}{1.00}

This paper proposes a network initialization method that obtains visual filters without using any real-world images.
The proposed method utilizes Perlin noise for initializing networks.
Perlin noise \cite{Perlin_1985} is a type of noise used in the field of computer graphics to render natural objects such as clouds, fire, and stones.
Specifically, the proposed method consists of two steps.
First, it generates Perlin noise samples with category labels defined based on noise complexity.
Second, it solves a classification problem, in which network parameters are optimized to classify the generated noise samples.
Compared with conventional initialization methods such as He initialization \cite{He_2015} based on Gaussian noise,
our method provides more complex texture-like visual filters as initial values.
We experimentally show that the proposed method outperforms 
conventional initialization methods utilized in most state-of-the-art image classification systems. Further, we demonstrate that the proposed method contributes to data-efficient pre-training in terms of accuracy improvement using fewer samples for pre-training.

In summary, this paper makes the following contributions:
\begin{enumerate}[leftmargin=28pt]
\item[(1)] A novel method is proposed for initializing networks, which solves an artificial classification problem without using real-world images. It consists of two components, namely a noise sample generator and a noise label generator, as shown in Figure~\ref{fig003}.
\item[(2)] Category definitions are given for Perlin noise samples based on noise complexity. This provides a specific definition of an artificial classification problem to be solved in the proposed method.
\item[(3)] Extensive experiments are conducted on four datasets. In addition to a performance comparison with conventional initialization methods, we show that the proposed method contributes to data-efficient pre-training.
\end{enumerate}

\section{Related Work}

This section summarizes the network architectures, initialization methods, pre-training methods, and datasets discussed in related studies.

\subsection{Network Architectures}
Many previous studies have shown the effectiveness of convolutional neural networks for image classification, which results from their ability to learn visual patterns from a large number of images.
The basic idea of convolution and pooling mechanisms was proposed in the 1990s \cite{LeCn_1998}. Since then, various types of network architectures have been proposed.
AlexNet \cite{Krizhevsky_2012}, which has seven hidden layers, was the first successful application of large-scale training using natural images.
VGGNet \cite{Simonyan_2015} and InceptionNet \cite{Szegedy_2015} stack more layers to explore deeper architectures.
ResNet \cite{He_2016} introduces skip connections to avoid the gradient vanishing problem.
ResNet also has extensions such as ResNeXt \cite{Xie_2017}, DenseNet \cite{Huang_2017}, and SE-ResNet \cite{Hu_2018},
and is the most widely used architecture for image classification tasks such as object recognition \cite{Russakovsky_2015}, action recognition \cite{Carreira_2017}, and scene understanding \cite{Zhou_2017}.

To train these networks, a large number of images are needed
because the networks typically have more than a million parameters to be optimized.
Therefore, training with a limited amount of data is a challenging problem in the field of pattern recognition and computer vision.

\subsection{Initialization Methods}
In most state-of-the-art optimization methods, the parameters of neural networks are initialized independently by utilizing probabilistic distributions such as the Gaussian distribution and the uniform distribution.

He initialization \cite{He_2015} is the most popular method, and utilizes a Gaussian distribution with a mean of zero and a variance that is scaled depending on the number of hidden units to initialize the weight parameters at each layer. This method is effective for recent deep convolutional networks with ReLU activations \cite{ICPR_Manessi_2018}.
Xavier initialization \cite{Xavier_2010} utilizes a uniform distribution to initialize weights. This is particularly effective for networks with a smooth activation function such as the sigmoid or tangent functions.
Sparse initialization \cite{Martens_2010} limits the number of non-zero initial weights. It was proposed with a Hessian-free optimization method.
Other classic initialization methods include normal initialization, which uses a standard normal distribution, and zero initialization, which assigns zero to all weights.
Zero initialization is a good choice for bias parameters or parameters that should be sparse.

Most of these methods can be viewed as initialization using Gaussian noise or uniform noise. In contrast, our idea is to utilize Perlin noise, a type of noise closer to natural visual patterns, for image classification.

\subsection{Pre-Training Methods}
Pre-training is a framework for initializing neural networks using a set of collected images.
The idea is to first train a network on a large-scale dataset, and then adopt the resulting network as the starting point for training on another dataset.

A recent trend for pre-training is to utilize a large-scale dataset such as ImageNet \cite{Russakovsky_2015}, Places~365 \cite{Zhou_2017}, and Kinetics~400-700 \cite{Carreira_2017,Kay_2017}.
Because these datasets consist of more than a million images with high-quality human-annotated labels, networks pre-trained on them have a reasonable set of visual filters at some hidden layers for recognizing natural objects from images.
This is one of the reasons why training from pre-trained parameters outperforms training from scratch in terms of image classification accuracy.
However, the use of these datasets is often limited to research and academic purposes.
Because it is not always easy to collect such a large number of images, data-efficient pre-training methods are desirable.

Our method is presented here as an initialization method because it does not use any natural images or any manually annotated labels. This also distinguishes our method from semi-supervised \cite{ICPR_Hailat_2018} \cite{ICPR_LChen_2018} \cite{ICPR_Ling_2018} \cite{ICPR_Robles_2016} and unsupervised \cite{ICPR_Ghaderi_2016} learning. Additionally, it can also be viewed as a new type of pre-training method because it has a step that solves a classification problem. Notably, we show experimentally that the proposed method contributes to data-efficient pre-training.

\subsection{Datasets}

Various datasets have been created for evaluating image classification methods.
Each dataset consists of a set of images and labels for a specific task, such as hand-written character recognition and object recognition.

Examples of hand-written character recognition datasets include MNIST \cite{LeCn_1998} and Omniglot \cite{Lake_2015}.
The MNIST dataset consists of 70,000 binary images of hand-written digits from 0 to 9.
The Omniglot dataset consists of about 40,000 grayscale images of 1,623 different hand-written characters from 50 different alphabets.
It was originally proposed as a dataset for one-shot learning to explore new learning methods.

Small- and large-scale object recognition datasets have been created.
Small-scale datasets are often used for evaluating image classification methods.
They include Cifar-10/100 \cite{Cifar10}, Caltech-101/256 \cite{FeiFei_2006}, and Pascal VOC \cite{voc}.
Each of these datasets consists of 10-100k natural images in 10-100 object categories.
The Describable Textures Dataset (DTD) \cite{Cimpoi_2014} focuses on texture categories related to objects.
Large-scale datasets such as ImageNet \cite{Russakovsky_2015} and COCO \cite{Lin_2014} are often used for pre-training.
They consist of more than a million natural images.

\fig{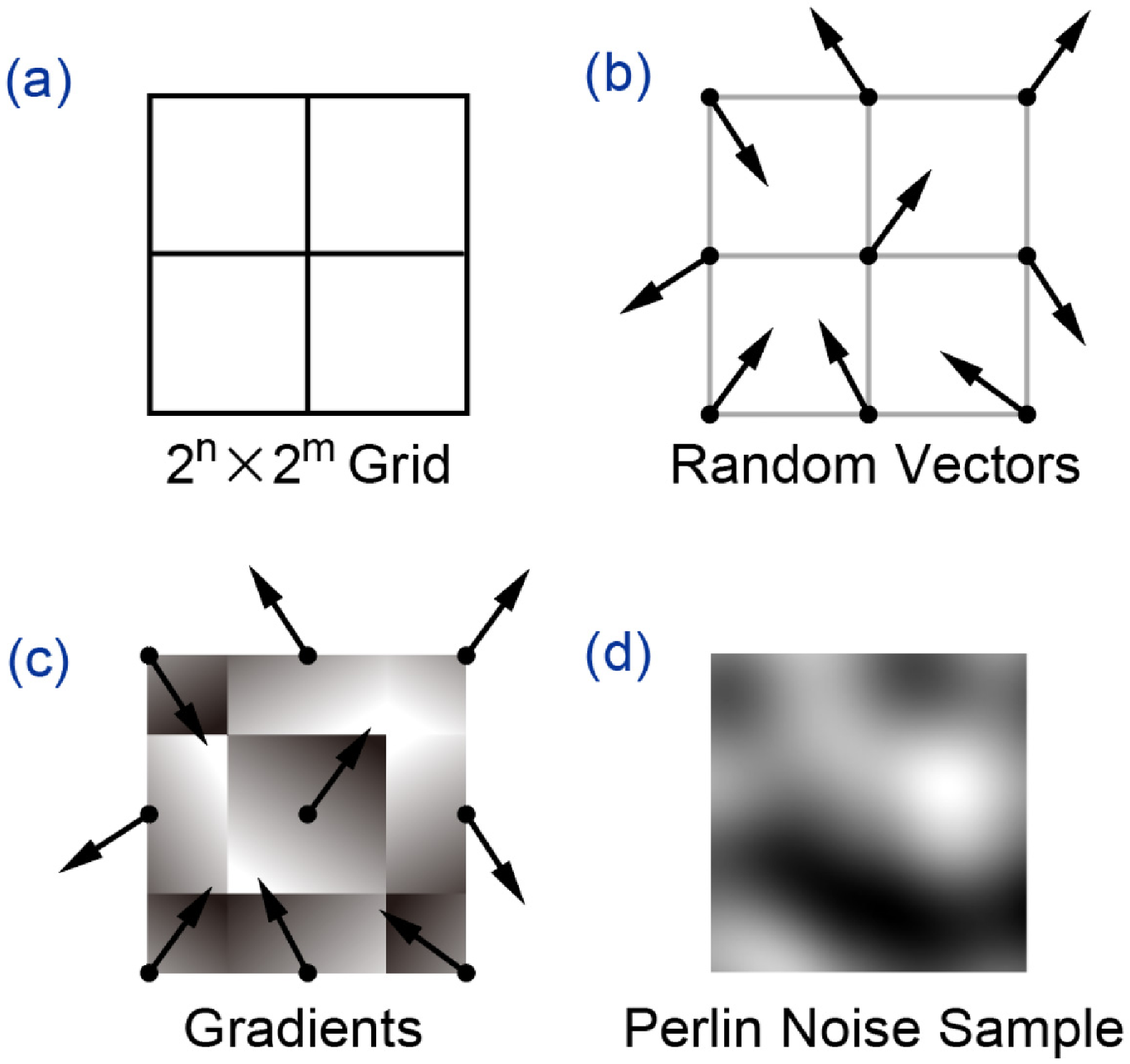}{Four steps used to generate Perlin noise. Noise samples are generated on a $2^{n} \times 2^{m}$ grid.\vspace{-6pt}}{1.0}

\section{Proposed Method}
In this section, we present the proposed network initialization method that uses Perlin noise. Let $\mathcal{N}_{\theta}$ be a neural network for image classification with a set of parameters $\theta$. The goal is to find reasonable initial values for $\theta$.

Our main idea is to initialize the parameters by solving an artificial noise classification problem. Specifically, the proposed method consists of two steps.
First, it generates noise data $\mathcal{D} = \{ (\epsilon_{i}, y_{i}) \}_{i=1}^{T}$, where $\epsilon_{i}$ is a Perlin noise sample and $y_{i}$ is a category label.
Second, it solves a classification problem on $\mathcal{D}$, in which network parameters are optimized to maximize the noise classification accuracy.
Note that even though this step can also be viewed as pre-training, the
entire process is proposed as an initialization method
because it does not use any natural images or any human-annotated labels.
To the best of our knowledge, this is the first work to initialize networks by solving an artificial classification problem using noise.
The rest of this section describes the details of each step.

\subsection{Generation of Noise Data}
This subsection describes the generation of noise data $\mathcal{D} = \{ (\epsilon_{i}, y_{i}) \}_{i=1}^{T}$ in the first step of the proposed method.
As shown in Figure~\ref{fig003},
the proposed method has two components, namely a noise sample generator $G_{\eta}$ and a noise label generator $L_{\eta}$ with shared parameter $\eta$.
Noise samples and their labels are generated as $\epsilon_{i} \sim G_{\eta}$ and $y_{i} \sim L_{\eta}$, respectively.

\fig{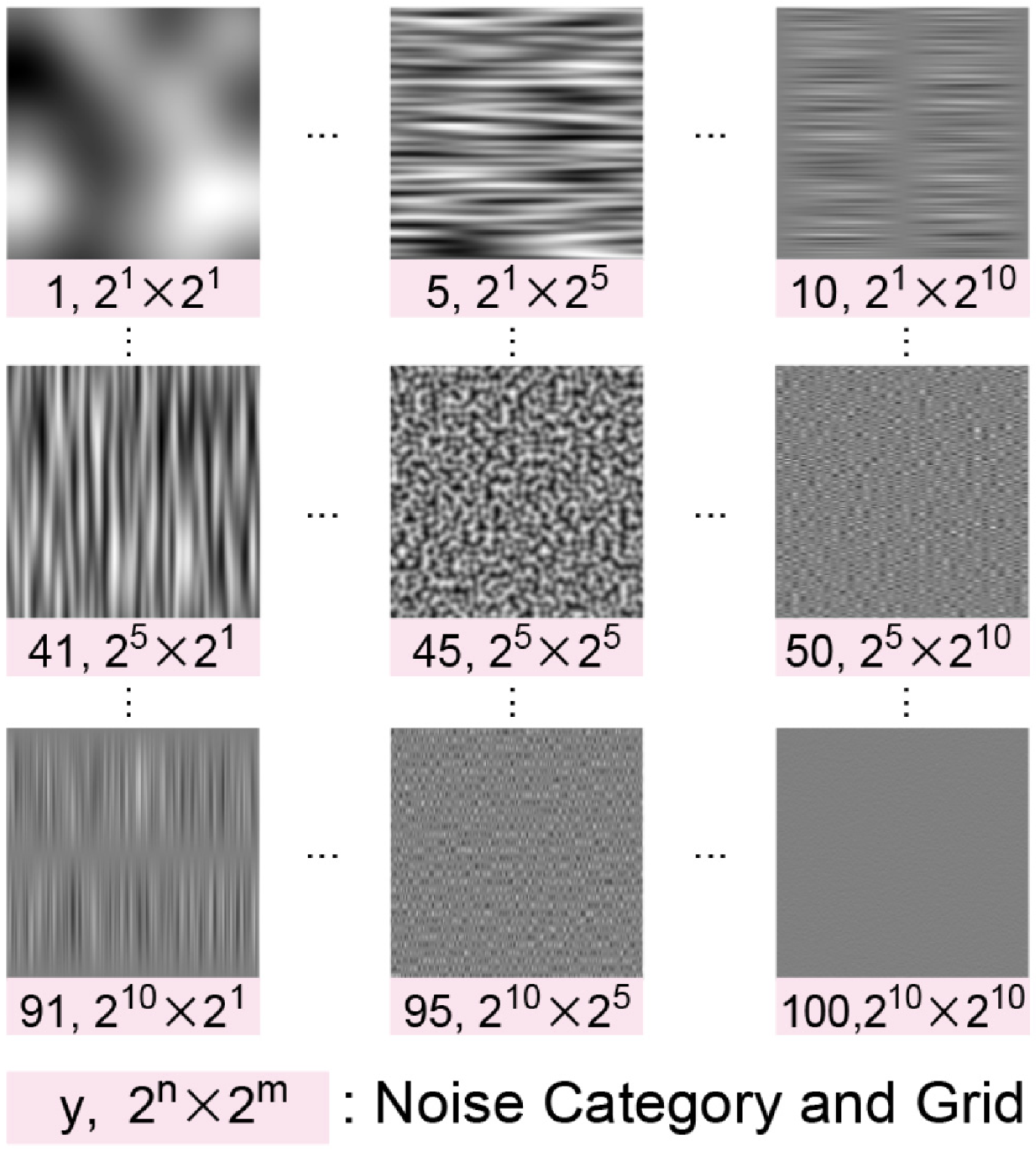}{Examples of noise categories. Labels attached to noise samples correspond to the complexity of Perlin noise. }{0.95}

In this work, we utilize Perlin noise \cite{Perlin_1985} to define these two generators.
Perlin noise was selected for the following reasons:
1) It can render the textures of some natural objects. For example, in the field of computer graphics, it is used to render clouds, fire, and stones.
2) It has moderate complexity for defining categories (see the visualization of some noise categories in Figure~\ref{fig002b}).
The definitions of the two generators using Perlin noise are given below.

\vspace{3pt}
\noindent{\it A-1. Noise Sample Generator}
\vspace{3pt}

Let $W, H$, and $C$ be the width, height, and number of channels of inputs of the network $\mathcal{N}_{\theta}$ for image classification, respectively.
The noise sample generator $G_{\eta}$ generates $\epsilon_{i} \in \mathbb{R}^{W \times H \times C}$ by following the Perlin noise generation algorithm in \cite{Perlin_1985}, summarized below. Note that it has parameter $\eta = (n, m)$ in Step~1 and random values in Step~2.

\noindent {\bf Step 1: Definition of a grid}.
This step defines a two-dimensional grid on a blank image whose size is $W \times H$. We define a $2^{n} \times 2^{m}$ grid ($n, m \geq 1$) as shown in Figure~\ref{fig002a}~(a).

\noindent {\bf Step 2: Placement of random gradient vectors}.
This step places random gradient vectors $v_{p,q}$ at each grid point, i.e.,
for $p=0,1,2,\cdots, 2^{n}$ and $q=0,1,2,\cdots, 2^{m}$, as shown in Figure~\ref{fig002a}~(b).
The magnitude and angle of each $v_{p,q}$ are uniformly sampled from $[0, R)$ and $[0, 2\pi)$, respectively. Here, we set $R = 0.01 \cdot \max(W,H)$.

\noindent {\bf Step 3: Generation of gradients}.
At each pixel, this step computes the dot product of
the gradient vectors at the four corners of the cell the pixel belongs to and the distance vectors between the pixel and the corresponding corners.
The four dot product values corresponding to the four corners are assigned to each pixel. This step is illustrated in Figure~\ref{fig002a}~(c). Note that the figure shows only one of the four values at each pixel.

\noindent {\bf Step 4: Interpolation}.
Finally, through linear interpolation between the four values computed in the previous step, the final value of each pixel on $\epsilon_{i}$ is determined. An example is shown in Figure~\ref{fig002a}~(d).

\vspace{3pt}
\noindent{\it A-2. Noise Label Generator}
\vspace{3pt}

In the above algorithm, the way that the interval gradient vectors are defined affects the complexity of the generated Perlin noise. 
For example, compared to noise samples computed from a dense grid, those computed from a sparser grid will also be sparser in terms of noise complexity.

We use this difference in complexity to define categories of noise samples.

Specifically, to samples $\epsilon_{i} \sim G_{\eta}$ generated with parameter $\eta = (n,m)$, the noise label generator $L_{\eta}$ attaches labels
\begin{align}
y_{i} = (n - 1) M + m.
\end{align}
Here, we assume that the range of $\eta$ is $\{(n,m): 1 \leq n \leq N,\ 1 \leq m \leq M\}$.
This means that the range of $y_{i}$ is $\{1,2,\cdots,NM\}$ and the number of noise categories is $NM$.

Figure~\ref{fig002b} and Figure~\ref{fig005} show examples of noise categories and intra-category variation of noise samples, respectively.
As can be seen, from category to category, noise complexity varies from coarse to fine with changes in gradient direction.
This artificial category definition facilitates
the acquisition of a reasonable set of visual filters
in a portion of the hidden layers of the neural network.
For example, with a convolutional neural network,
we obtain some texture-like filters
at the bottom layer of the network (results are shown in Figure~\ref{fig004} in Sec.~4).

Note that the proposed method has three hyperparameters, namely $N$, $M$, and $K$.
$N$ and $M$ are for determining the number of noise categories and
$K$ is the number of instances per category.
Finally, the generated noise data $\mathcal{D}$ consist of $T=NMK$ noise samples with labels.

\fig{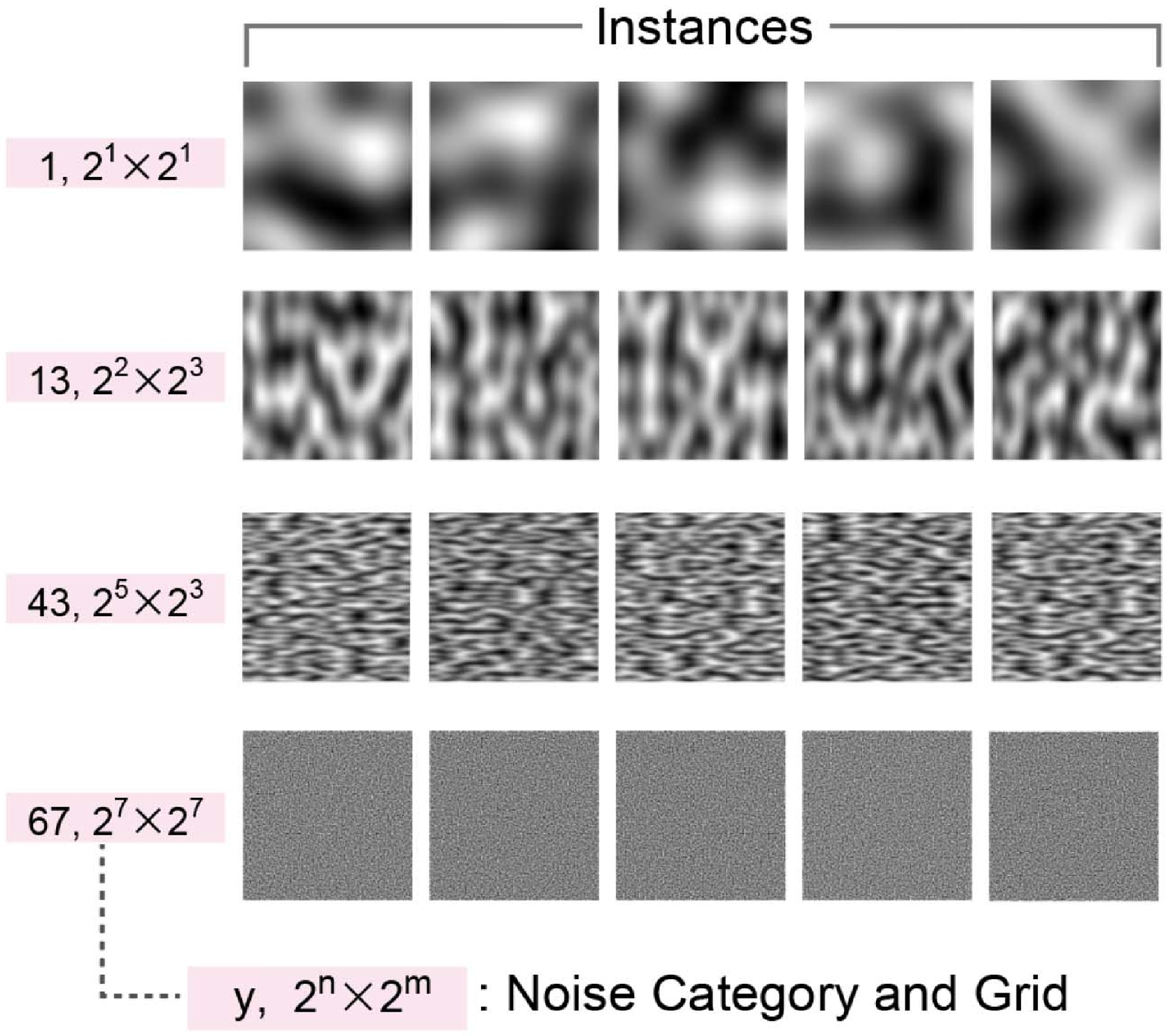}{Intra-category variation of noise samples. Image representations of noise samples for three categories ($y=1, 13, 43, 67$) are shown.}{1.0}

\tableA
\figw{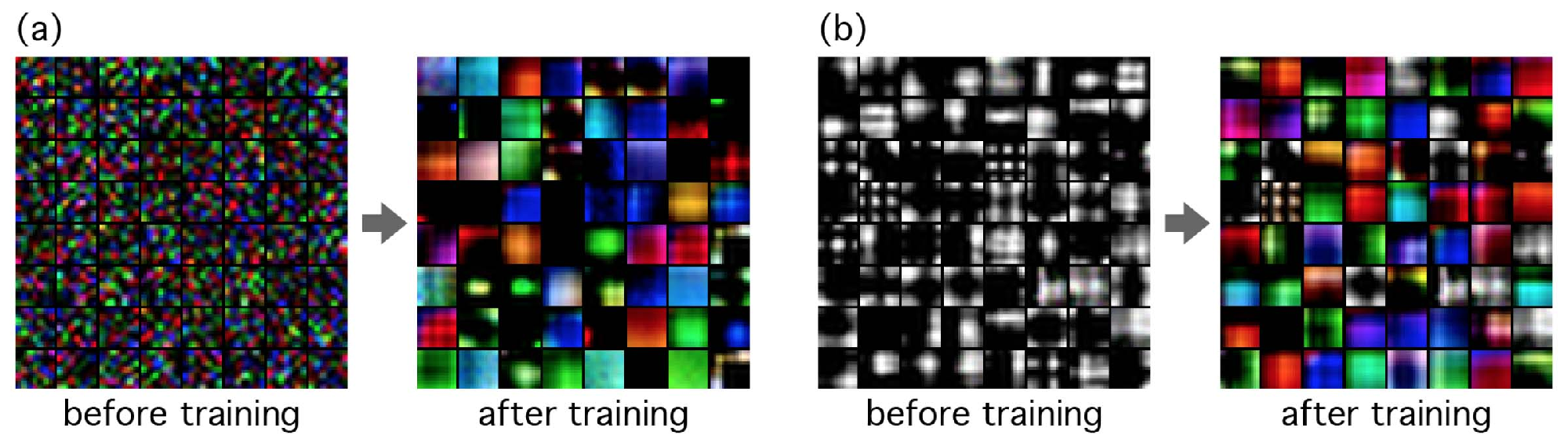}{Visualization of filters of the first convolutional (conv1) layer. Filters before and after training on Cifar-10 are shown. (a) He initialization and (b) proposed method.}{1.0}

\subsection{Optimization}
The network parameters are optimized by solving a classification problem on the generated noise data $\mathcal{D} = \{ (\epsilon_{i}, y_{i}) \}_{i=1}^{T}$.
In this step, any type of objective function and optimizer can be introduced.
Examples include cross-entropy loss with the SGD or ADAM optimizer, which starts from He initialization.
In contrast to standard pre-training, which solves an optimization problem on real-world data,
this step solves an artificial optimization problem,
and as such this is a kind of closed-form problem.
Introducing other types of optimizers to this step to solve this closed-form problem more efficiently, based on the characteristics of Perlin noise, will be considered in future work.

\section{Experiments}
In this section, we show the effectiveness of the proposed initialization method on four image classification datasets.

\subsection{Datasets and Evaluation Measures}

\noindent {\bf Cifar-10}.
This dataset consists of 60,000 color images,
each of which has a label from 10 object classes.
We follow the standard evaluation procedure, in which 50,000 images are used for training and 10,000 images are used for testing.
Classification accuracy over 10 classes is reported.

\noindent {\bf Cifar-100}.
This dataset consists of 60,000 color images,
each of which has a label from 100 object classes.
We follow the standard evaluation procedure, in which 500 and 100 images per class are used for training and testing, respectively.
Classification accuracy over 100 classes is reported.

\noindent {\bf Omniglot}.
This dataset consists of 38,300 grayscale images of 1,623 different hand-written characters. It was originally proposed as a one-shot learning dataset.
We use the data split proposed in \cite{Lake_2015}.
1,623 and 30,837 images are used for training and testing, respectively.
Classification accuracy over 1,623 classes is reported.

\noindent {\bf Describable Textures Dataset (DTD)}.
This dataset consists of 5640 images of 47 texture categories, such as checkered, striped, meshed, and marbled.
We follow the standard evaluation procedure with three equal parts for training, validation, and testing.

\noindent {\bf ImageNet}.
This dataset consists of 1.2 million color images of 1,000 object classes.
We use this dataset to show that the proposed initialization method
helps improve the data efficiency of pre-training.

\subsection{Implementation Details}
Our implementation uses ResNet \cite{He_2016} as a backbone network.
The results obtained using ResNet50 and ResNet152 are reported.
For optimization, we use cross-entropy loss and the SGD optimizer in all experiments.
For comparison, we report results obtained using
He initialization \cite{He_2015}, Xavier initialization \cite{Xavier_2010}, sparse initialization \cite{Martens_2010}, and normal initialization.
Notably, He initialization is utilized in most state-of-the-art methods for image classification tasks.

\subsection{Results}

\noindent{\it C-1. Main Results}

Table~\ref{tableA} shows a performance comparison of initialization methods on the four datasets.
As shown, the proposed method always outperforms the conventional methods.
This shows the effectiveness of initialization using Perlin noise.

To analyze the reason for the superior performance of the proposed method,
Figure~\ref{fig004} shows the filters at the bottom layer (conv1 layer) before and after training.
As can be seen, the proposed method gives a set of various filters, including some texture-like and edge-like filters.
Therefore, after training,
it obtains a useful set of colored filters for image classification even if the number of training samples is not very large.

Figure~\ref{fig006} shows validation accuracy curves for the Cifar-100 dataset. Our method not only achieves higher accuracy at the end of training, but its training starts at higher accuracy. This shows that, as an initial value for training, texture and edge filters are a better choice than randomized filters. In this experiment, we used the fixed learning rate schedule in the official PyTorch implementation tuned for training from scratch for a fair comparison. Optimizing the learning rate schedule for our method may lead to faster convergence. 

In the present paper, we did not use color Perlin noise because
some applications, including hand-written character recognition, use grayscale images.
In future work, a noise sample generator that uses the RGB color space can be applied for cases where the dataset consists of color images.

\ \\\noindent{\it C-2. Hyperparameters}

To explore the effect of changes in hyperparameters, Table~\ref{tableB} shows the results obtained using various values for $N$, $M$, and $K$ for noise data generation. Note that, as described in Sec.~3, $(N,M)$ determines the number of Perlin noise categories based on noise complexity, and $K$ controls the number of instances per category for generating $T=NMK$ noise samples.

\fig{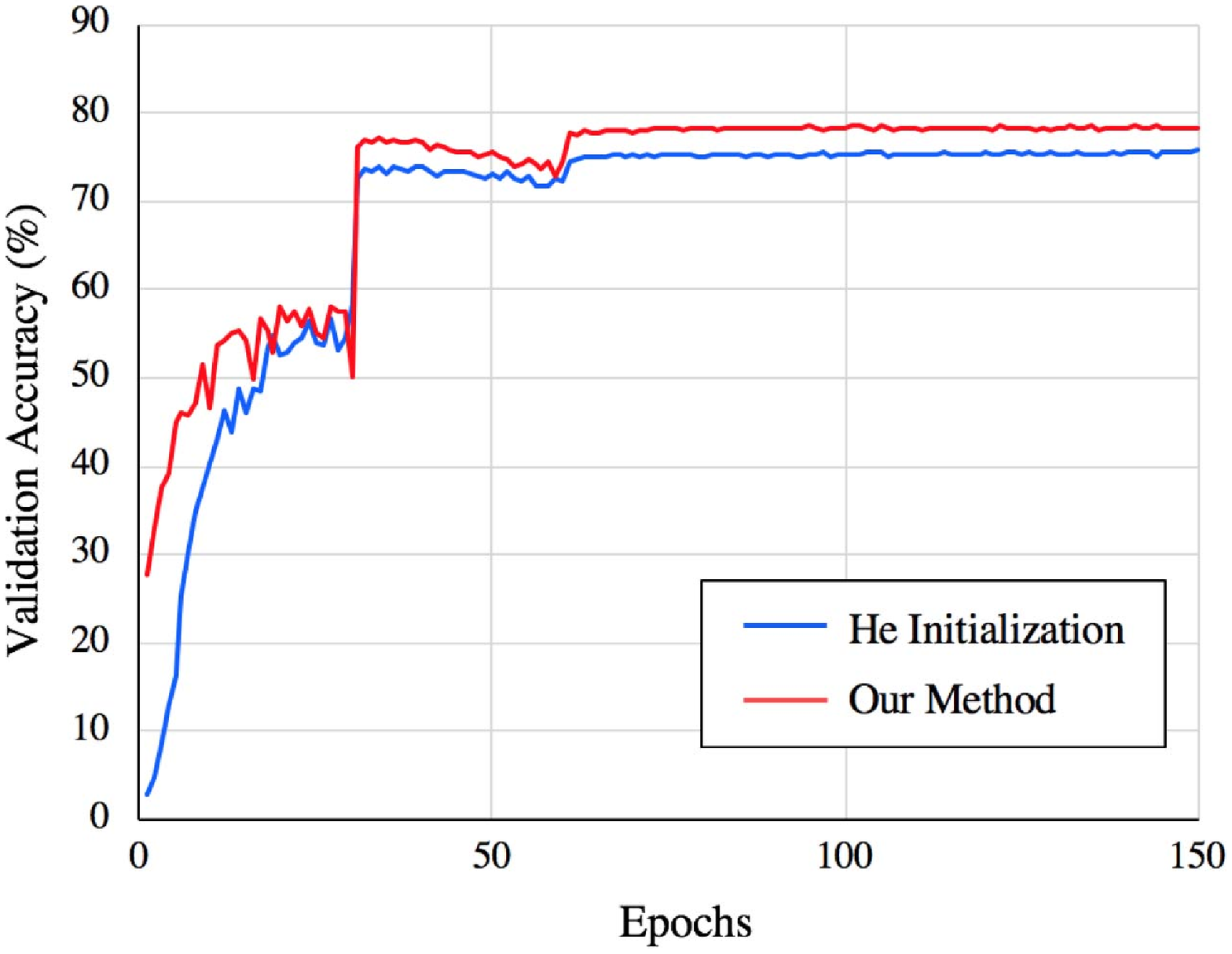}{Validation accuracy curves for the Cifar-100 dataset. He initialization and our method are compared.\vspace{-7pt}}{0.95}

A comparison of two strategies, namely increasing the number of categories and increasing the number of instances, shows that the former is more effective than the latter for improving image classification accuracy. This supports our assumption that noise complexity is an important factor for determining categories.

Performance improvements were observed across both network structures, with ResNet152 always outperforming ResNet50.
In general, it is difficult to train a large network with a limited amount of data.
However, our method is adaptable to networks of different sizes
because noise complexity can be adjusted based on the network size.
This advantage improves the performance of training large networks with a limited amount of data.
To further improve performance, 
future noise sample generators can be designed using other types of noise.

\ \\\noindent{\it C-3. Data-Efficient Pre-Training}

Finally, we combine our method with pre-training to explore how this combination improves image classification performance.
Figure~\ref{fig007} shows the results obtained using various numbers of images from the ImageNet (ILSVRC 2012) dataset for pre-training.
According to the results, our method is particularly effective when the number of pre-training images is small.
Although this paper focuses on network initialization without using any real-world images, the results imply that combining our method with pre-training methods would lead to more data-efficient pre-training.

\tableB

\section{Conclusion}
This paper proposed a novel network initialization method that uses Perlin noise.
Experiments
on four image classification datasets demonstrated that the proposed initialization is effective for training networks for image classification.
In particular, our method is effective for training with a limited amount of data because it provides texture-like initial filters that improve image classification accuracy.
Our future work will focus on extending this initialization method to other types of noise, including colored noise, and
data such as audio and video data, as well as introducing new noise category definitions.

\fig{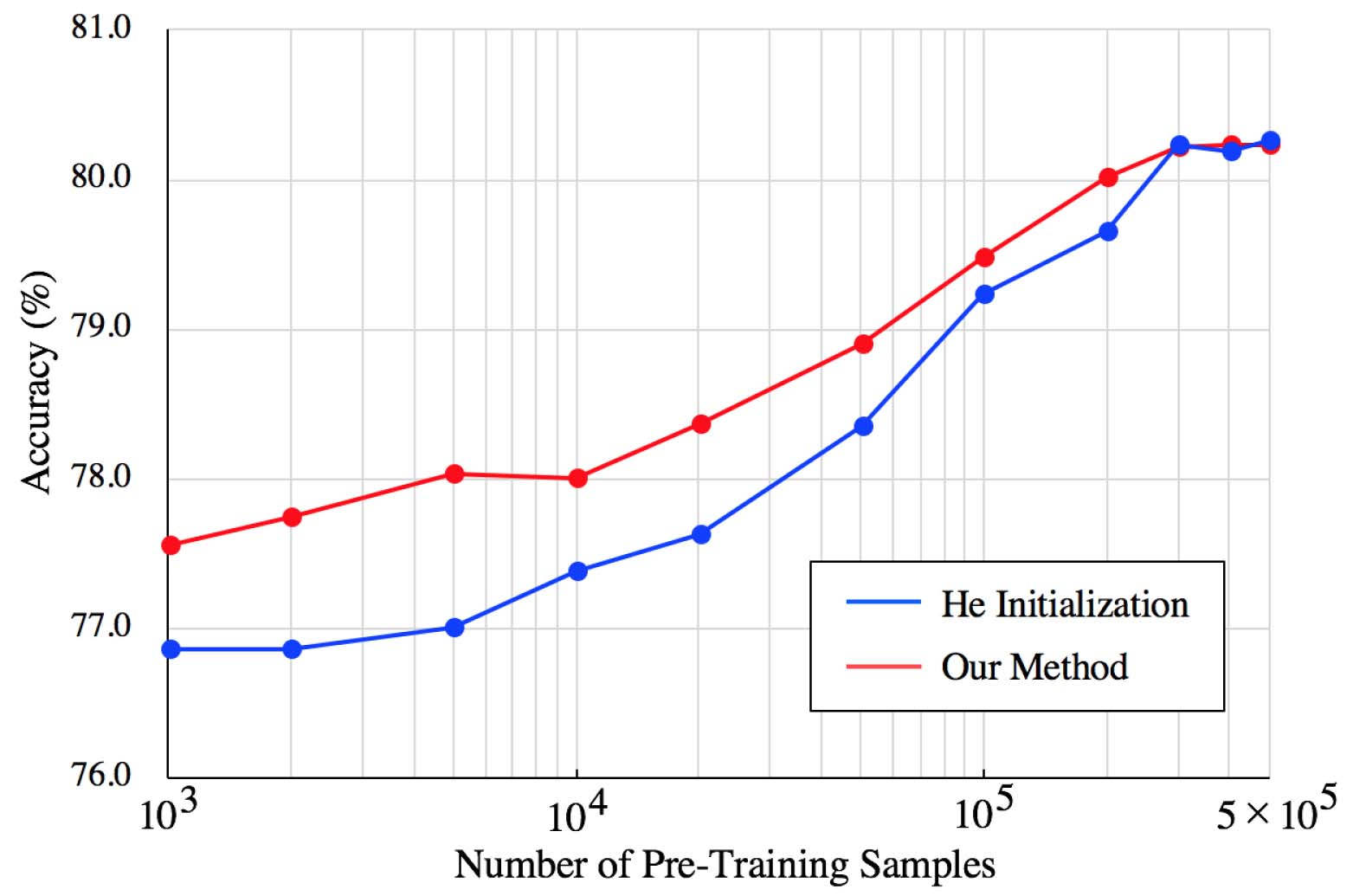}{Performance comparison using various numbers of samples for pre-training. Accuracy for the Cifar-100 dataset is shown. ResNet152 was pre-trained using images from the ImageNet dataset.
}{0.98}

\section*{Acknowledgment}
This work was partially supported by grants JSPS KAKEN 19H01134 and JST ACT-I (JPMJPR16U5). Computational resource of AI Bridging Cloud Infrastructure (ABCI) provided by National Institute of Advanced Industrial Science and Technology (AIST) was used.


\begin{thebibliography}{99}
\bibitem{Russakovsky_2015}
O.~Russakovsky, \etal{J.~Deng, H.~Su, J.~Krause, S.~Satheesh, S.~Ma, Z.~Huang, A.~Karpathy, A.~Khosla, M.~Bernstein, A.C.~Berg, and L.~Fei-Fei.}
ImageNet Large Scale Visual Recognition Challenge.
{\jornal Springer IJCV}, vol.~115, no.~3, pp.~211--252, 2015.

\bibitem{Perlin_1985}
K.~Perlin,
An Image Synthesizer.
{\jornal SIGGRAPH}, pp.~287--296, 1985.


\bibitem{He_2015}
K.~He, \etal{X.~Zhang, S.~Ren, and J.~Sun.}
Delving Deep into Rectifiers: Surpassing Human-Level Performance on ImageNet Classification.
{\proc ICCV}, pp.~1026--1034,~2015.

\bibitem{LeCn_1998}
Y.~LeCun, \etal{L.~Bottou, Y.~Bengio, and P.~Haffner.}
Gradient-based Learning Applied to Document Recognition.
{\em Proc. of the IEEE}, pp.~2278--2324, 1998.

\bibitem{Krizhevsky_2012}
A.~Krizhevsky, \etal{I.~Sutskever, and G.E.~Hinton.}
ImageNet Classification with Deep Convolutional Neural Networks.
{\proc NeurIPS}, pp.1--9, 2012.

\bibitem{Simonyan_2015}
K.~Simonyan and A.~Zisserman.
Very Deep Convolutional Networks for Large-Scale Image Recognition.
{\proc ICLR}, 2015.

\bibitem{Szegedy_2015}
C.~Szegedy, \etal{W~Liu, Y~Jia, P.~Sermanet, S.~Reed, D.~Anguelov, D.~Erhan, V.~Vanhoucke and A.~Rabinovich.}
Going Deeper with Convolutions.
{\proc  CVPR}, 2015.

\bibitem{He_2016}
K.~He, \etal{X.~Zhang, S.~Ren and J.~Sun.}
Deep Residual Learning for Image Recognition.
{\proc CVPR}, 2016.

\bibitem{Xie_2017}
S.~Xie, \etal{R.~Girshick, P.~Dollar, Z.~Tu, and K.~He.}
Aggregated Residual Transformations for Deep Neural Networks.
{\proc CVPR}, 2017.

\bibitem{Huang_2017}
G.~Huang, \etal{Z.~Liu, L.~van~der~Maaten, and K.~Q.~Weinberger.}
Densely Connected Convolutional Networks.
{\proc CVPR}, 2017.

\bibitem{Hu_2018}
J.~Hu, \etal{L.~Shen, S.~Albanie, G.~Sun, and E.~Wu.}
Squeeze-and-Excitation Networks.
{\proc CVPR}, 2018.

\bibitem{Carreira_2017}
J.~Carreira and A.~Zisserman.
Quo Vadis, Action Recognition? A New Model and the Kinetics Dataset.
{\proc CVPR}, 2017.

\bibitem{Zhou_2017}
B.~Zhou, \etal{A.~Lapedriza, A.~Khosla, A.~Oliva, and A.~Torralba.}
Places: A 10 million Image Database for Scene Recognition.
{\jornal IEEE Trans. on PAMI}, vol.~40, no.~6, pp.~1452--1464, 2017.

\bibitem{ICPR_Manessi_2018}
F. Manessi and A. Rozza. Learning Combinations of Activation Functions. {\proc ICPR}, pp. 61--66, 2018.


\bibitem{Xavier_2010}
X.~Glorot and Y.~Bengio.
Understanding the Difficulty of Training deep Feedforward Neural Networks.
{\proc AISTAS}, pp.~249--256, 2010.

\bibitem{Martens_2010}
J.~Martens.
Deep Learning via Hessian-free Optimization.
{\em Proc. ICML}, 2010.

\bibitem{Kay_2017}
W.~Kay, \etal{J.~Carreira, K.~Simonyan, B.~Zhang, C.~Hillier, S.~Vijayanarasimhan, F.~Viola, T.~Green, T.~Back, P.~Natsev, M.~Suleyman, and A.Zisserman.}
The Kinetics Human Action Video Dataset.
{\em CoRR 1705.06950}, 2017.

\bibitem{ICPR_Hailat_2018}
Z. Hailat, \etal{A. Komarichev and X. Chen,} Deep Semi-Supervised Learning. {\proc ICPR}, pp. 2154--2159, 2018.

\bibitem{ICPR_LChen_2018}
L. Chen, \etal{S. Yu and M. Yang,} Semi-Supervised Convolutional Neural Networks with Label Propagation for Image Classification. {\proc ICPR}, pp.~1319--1324, 2018.

\bibitem{ICPR_Ling_2018}
Z. Ling, \etal{X. Li, W. Zou and S. Guo,} Semi-Supervised Learning via Convolutional Neural Network for Hyperspectral Image Classification. {\proc ICPR}, pp.~1--6,~2018.

\bibitem{ICPR_Robles_2016}
A. Robles-Kelly and Ran Wei. Semi-Supervised Image Labelling Using Barycentric Graph Embeddings. {\proc ICPR}, pp. 1518--1523, 2016.


\bibitem{ICPR_Ghaderi_2016}
A. Ghaderi and V. Athitsos. Selective Unsupervised Feature Learning with Convolutional Neural Network. {\proc ICPR}, pp. 2486--2490, 2016.


\bibitem{Lake_2015}
B.~M.~Lake, \etal{R.~Salakhutdinov, and J.B.~Tenenbaum.}
Human-level Concept Learning Through Probabilistic Program Induction.
Science, vol.~350(6266), pp.~1332--1338, 2015.

\bibitem{Cifar10}
A.~Krizhevsky.
Learning Multiple Layers of Features from Tiny Images.
Technical Report TR-2009, University of Toronto, 2019.

\bibitem{FeiFei_2006}
L.~Fei-Fei, \etal{R.~Fergus and P.~Perona.}
One-Shot learning of object categories.
{\em IEEE Trans. Pattern Recognition and Machine Intelligence}, vol.~28 , no.~4, 2006.

\bibitem{voc}
M.~Everingham, \etal{S.M.A.~Eslami, L.V.~Gool, C.K.I.~Williams, J.~Winn, and A.~Zisserman.}
The Pascal Visual Object Classes (VOC) Challenge.
{\jornal Springer IJCV}, vol.~88, no.~2, pp.~303--338, 2010.

\bibitem{Cimpoi_2014}
M.~Cimpoi, \etal{S. Maji, I. Kokkinos, S. Mohamed, and A. Vedaldi.}
Describing Textures in the Wild.
{\proc CVPR}, 2014.

\bibitem{Lin_2014}
T.-Y.~Lin, \etal{M.~Maire, S.~Belongie, J.~Hays, P.~Perona, D.~Ramanan, P.~Dollar, and C.L.~Zitnick.}
Microsoft COCO: Common Objects in Context.
{\proc ECCV}, 2014.




\end{thebibliography}
\end{document}